\documentclass[conference]{IEEEtran}

\usepackage{cite}
\usepackage{amsmath,amssymb,amsfonts}
\usepackage{algorithmic}
\usepackage{graphicx}
\usepackage{textcomp}
\usepackage{xcolor}

\usepackage{comment}
\usepackage{dsfont}
\usepackage{float}
\usepackage{soul,color}
\usepackage{hyperref}
\usepackage{physics}
\usepackage{mathtools}
\usepackage[numbers]{natbib} 
\usepackage{flushend} 

\title{\huge Compositional Factorization of Visual Scenes with Convolutional Sparse Coding and Resonator Networks}

\begin{document}

\author{\IEEEauthorblockN{Christopher J. Kymn\IEEEauthorrefmark{1}\IEEEauthorrefmark{5},
Sonia Mazelet\IEEEauthorrefmark{1}\IEEEauthorrefmark{2}\IEEEauthorrefmark{5},
Annabel Ng\IEEEauthorrefmark{1}, 
Denis Kleyko\IEEEauthorrefmark{3}\IEEEauthorrefmark{4} and
Bruno A. Olshausen\IEEEauthorrefmark{1}}
\IEEEauthorblockA{\IEEEauthorrefmark{1}Redwood Center for Theoretical Neuroscience, UC Berkeley,
Berkeley, USA}
\IEEEauthorblockA{\IEEEauthorrefmark{2}Université Paris-Saclay, ENS Paris-Saclay, Gif-sur-Yvette, France\\}
\IEEEauthorblockA{\IEEEauthorrefmark{3}Centre for Applied Autonomous Sensor Systems, \"{O}rebro University, \"{O}rebro, Sweden\\}
\IEEEauthorblockA{\IEEEauthorrefmark{4}Intelligent Systems Lab, Research Institutes of Sweden, Kista, Sweden\\}
\IEEEauthorblockA{\IEEEauthorrefmark{5}Equal contribution\\}
}

\maketitle
\begin{abstract}
We propose a system for visual scene analysis and recognition based on encoding the sparse, latent feature-representation of an image into a high-dimensional vector that is subsequently factorized to parse scene content. 
The sparse feature representation is learned from image statistics via convolutional sparse coding, while scene parsing is performed by a {\em resonator network} \cite{FradyResonator2020}. 
The integration of sparse coding with the resonator network increases the capacity of distributed representations and reduces collisions in the combinatorial search space during factorization. We find that for this problem the resonator network is capable of fast and accurate vector factorization, and we develop a confidence-based metric that assists in tracking the convergence of the resonator network. 
\end{abstract}

\begin{IEEEkeywords}
hyperdimensional computing, vector symbolic architectures, sparse coding, resonator networks, visual scene understanding, combinatorial search, vector factorization, computing-in-superposition
\end{IEEEkeywords}

\section{Introduction}

Visual perception requires making sense of previously unencountered combinations of objects and their poses in a visual scene.  
Although this process can feel effortless, we still lack a 
concrete and universally
accepted computational model for how such strong generalization is achieved. Recent efforts in computer vision, including deep learning models for object recognition, typically require thousands of examples to learn an object class, yet remain both excessively sensitive to small perturbations in image pixel values \cite{goodfellow2015explaining} and excessively invariant to large semantic changes \cite{jacobsen2018excessive}. These instances suggest a divergence from the strategies used by humans, who can learn and robustly recognize new visual categories from limited examples \cite{morgenstern2019one}.

Learning new objects from limited examples is possible when one exploits the idea of compositionality. For example, a person who has visual concepts for ``blue'' and ``rose'' could recognize a blue rose upon seeing one, even if they have not explicitly encountered that conjunction. Though compositionality allows generalization, we still need a computational framework describing the basic compositional primitives, the operations by which they are combined, and how they are disentangled from pixel values in an image. For example, in vision, we  need to understand how the concepts ``blue'' and ``rose'' are represented by neural populations, how they are bound together to represent the image of a blue rose, and how such a representation is factorized apart into color and (floral) identity.  Though such problems of composition and factorization abound in perception, we currently lack a mechanistic model of how they are performed by the brain.

Hyperdimensional computing~\cite{kanerva2009hyperdimensional}, also known as vector symbolic architectures~\cite{gayler1998multiplicative} (HD/VSA), provides an explicit framework for modeling compositional structure and for manipulating and factorizing compositional representations~\cite{kleyko2022vector}. With HD/VSA, data structures are represented as random high-dimensional vectors of dimension $D$ typically between $1{,}000$ and $10{,}000$. Here we consider vectors where each component is a complex number with unit amplitude. This HD/VSA model, also known as Fourier Holographic Reduced Representation \cite{plate1995holographic}, is equipped with the component-wise sum (+) for bundling items together, the Hadamard product (component-wise multiplication, $\odot$) for binding items together, and the inner product operation ($\langle \cdot, \cdot  \rangle$) for measuring similarity between vectors. Under this framework, compositional representations are formed by binding and bundling vectors (for example, to represent a set of role-filler pairs). However, it is also necessary to invert this operation -- i.e., to decompose a compound structure (represented as a vector) to recover the individual elements (also vectors) that were combined to generate it. This amounts to a vector factorization problem.

The resonator network~\cite{FradyResonator2020} has been proposed as an efficient solution to the vector factorization problem within HD/VSA, and it has recently been successfully applied to the problem of disentangling objects and their poses in a visual scene~\cite{renner2022neuromorphic}.  
That approach utilized a direct encoding of image pixel values into a high-dimensional vector, as a first proof of concept.
However, an issue that arises in extending this approach to natural images is the non-orthogonality of vectors that arises due to strong correlations among image pixel values \cite{field1987relations}. To address this gap, we propose a solution based on first forming a sparse representation of the image using convolutional sparse coding~\cite{zeiler2010deconvolutional}, which learns a dictionary of elementary patterns contained in images. We show that the proposed approach produces closer-to-orthogonal vector representations and performs more accurately and efficiently than previous methods or each step individually, enabling applications of the resonator network for more difficult kinds of factorization problems.

\section{Methods}

The problem we seek to solve can be stated as follows: given an image containing one or more  objects, can we return each object's identity and position in the image? To address this problem, we take three steps: 1) infer a latent representation of the image with convolutional sparse coding, 2) convert the latent representation into a vector that can be factorized into each object and its pose, and 3) factorize the high-dimensional vector with the resonator network. We describe each step of this procedure in detail below.

\subsection{Learning and inferring latent image representations with convolutional sparse coding}

In designing useful latent representations of images for factorization, two criteria are paramount. First, the image representation should \textit{reduce redundancy}, i.e., minimize statistical dependencies among elements of the representation, so as to make explicit the structure contained in the data. Second, the image representation should be \textit{equivariant} to common classes of image transformations. In this case, we are concerned with translations: if the image representation is translated by some $2$D shift, there should likewise be a corresponding shift in the latent representation.

Convolutional sparse coding~\cite{zeiler2010deconvolutional} is an unsupervised learning algorithm that satisfies both of these properties. For each $2$D image $\mathbf{I}$, we infer a set of latent feature maps, $\{\mathbf{A}_j\}$. This latent representation expresses the image as a linear combination of convolved basis functions $\boldsymbol\phi_j$ via the following equation:

\begin{equation}
    \mathbf{I} = \sum_j [ \mathbf{A}_j \ast \boldsymbol\phi_j] + \mathcal{N},
\end{equation}

\noindent where $\ast$ denotes convolution, and $\mathcal{N}$ expresses residual structure (assumed to be Gaussian) not accounted for by the basis functions. This is a generalization of the original formulation of sparse coding~\cite{olshausen1996emergence}, which was defined for local regions of images. In convolutional sparse coding, both the basis functions $ \{ \boldsymbol\phi_j \}$, and latent feature maps $\{\mathbf{A_j}\}$, are optimized by minimizing the following objective function:

\begin{equation}
    E = \frac{1}{2}|| \mathbf{I} - \sum_{j=1}^{n}\boldsymbol\phi_j \ast \mathbf{A}_j ||^2_2 + \lambda \sum_{j=1}^{n}||\mathbf{A}_j||_1
\end{equation}
\noindent
where $\mathbf{I}$ is an image of size $L\times L$, $\boldsymbol\phi_j$ is the $j$-th learned basis function, $\mathbf{A}_j$ is a sparse feature map of size $L\times L$, and $\lambda$ is a regularization hyperparameter controlling the tradeoff between the first and second terms. The basis functions $\{ \boldsymbol\phi_j \}$ are optimized over an entire dataset of images, while the feature maps $\{ \mathbf{A}_j \}$ are inferred for each particular example given the fixed~$\{ \boldsymbol\phi_j \}$.

\subsection{Encoding the sparse representation into an HD/VSA vector}

Next, we transform $\mathbf{A}$ into a dense, high-dimensional vector $\mathbf{z(A)} \in \mathbb{C}^D$. As we will show, this transformation allows us to express the image factorization problem as a vector factorization problem, enabling subsequent inference via the resonator network. More specifically, we use Vector Function Architecture (VFA)~\cite{frady2021computing, fradyfunctionsnice2022}, a generalization of HD/VSA to continuous-valued variables and functions, and show that this mathematical framework enables use of the resonator network for finding objects and their transformations.

First, we generate random $D$-dimensional vectors $\mathbf{h}(x),\mathbf{v}(y),\mathbf{b}(j)$ for distinct horizontal positions (indexed by $x$), vertical positions (indexed by $y$), and basis functions (indexed by $j$). Each vector consists of phasors, i.e., $e^{i\omega}$ with $i = \sqrt{-1}$ and $\omega \in [- \pi , \pi )$. Each $\mathbf{b}(j)$ is chosen i.i.d. with phases chosen randomly from a uniform distribution, while $\mathbf{h}(x)$ and $\mathbf{v}(y)$ are selected according to fractional power encoding~\cite{PlateRecurrent1992} with periodic kernels~\cite{kymn2023computing}. That is, we first define a basis vector $\mathbf{h} = \mathbf{h}(1) = [e^{i\omega_1},\dots,e^{i\omega_D}]$, and then define $\mathbf{h}(x) = [e^{i\omega_1 x},\dots,e^{i\omega_D x}]$. Each phase $\omega$ is randomly chosen from the $L$-th roots of unity, ensuring that the representation is periodic with period $L$. This condition ensures that image transformations are invertible, which would not happen if images could disappear over the boundary.

We then form an image representation as follows:

\begin{equation}
    \mathbf{z(A)} =  \sum_{x,y,j} A_j(x,y) \cdot \mathbf{h}(x) \odot \mathbf{v}(y) \odot \mathbf{b}(j)
\end{equation}

\noindent where $\cdot$ denotes scalar-vector multiplication, and $\odot$ denotes component-wise vector multiplication (Hadamard product). We write the image encoding in this format to express it in terms of the bundling and binding operations of HD/VSA. 

Finally, we generate high-dimensional vector representations, $\mathbf{o}^{(k)}$ of each object ($K$ total objects, indexed by $k$) by calculating its convolutional sparse code for that object placed within a \textit{canonical reference frame}, and converting it to an HD/VSA vector in the format above. This allows us to represent scenes with one object as a binding of the high-dimensional vector of a canonical object $k'$ and the corresponding shifts $(x', y')$:

\begin{equation}
    \mathbf{q} = \mathbf{h}(x') \odot \mathbf{v}(y') \odot \mathbf{o}^{(k')}.
\end{equation}
\noindent
Our goal, then, is to infer the latent parameters $x',y',k'$, and we accomplish this with the resonator network.

\subsection{Factorizing the scene vector with a resonator network}

The resonator network~\cite{FradyResonator2020} is an algorithm for factoring a composition of bound vectors into the individual elements that composed it. To make the problem well posed, it is assumed that the maximum number of factors is known, and that each factor comes from a discrete set of possibilities. Even under these assumptions, the problem remains difficult due to the size of the search space. In our problem setting, the brute force approach for the factorization problem requires $L \times L \times K$ comparisons. By contrast, the resonator network solves the factorization problem by \textit{searching in superposition}; that is, it recursively updates weighted combinations of estimated factors until convergence. It has been shown empirically that this strategy converges faster and more accurately than other methods, including brute force search and gradient-based methods, on vector factorization problems~\cite{kent2020resonator,kymn2023computing}. 

Given a $D$-dimensional vector $ \mathbf{q} = \mathbf{h}(x') \odot \mathbf{v}(y') \odot \mathbf{o}^{(k')}$, the resonator network recovers the factors $\mathbf{h}(x'), \mathbf{v}(y'), \mathbf{o}^{(k')}$ among the finite number of possibilities $(\mathbf{h}(0),...,\mathbf{h}(L-1))$, $(\mathbf{v}(0),...,\mathbf{v}(L-1))$, and $(\mathbf{o}^{(1)},...,\mathbf{o}^{(K)})$. Vectors corresponding to the possible values of $\mathbf{h}$, $\mathbf{v}$, and $\mathbf{o}$ are stored in codebooks, i.e., matrices $\mathbf{H}=\{\mathbf{h}(0),...,\mathbf{h}(L-1)\}$, $\mathbf{V}=\{\mathbf{v}(0),...,\mathbf{v}(L-1)\}$, and $\mathbf{O}=\{\mathbf{o}^{(1)},...,\mathbf{o}^{(K)}
\}$ of sizes $D\times L$, $D\times L$ and $D\times K$, respectively.

The resonator network dynamics are defined by the following set of equations:

\begin{equation}\label{eq:resonator}
\begin{split}
    \hat{\mathbf{h}}_{t+1}=g(\mathbf{H}\mathbf{H}^\dag( \mathbf{z} \odot \hat{\mathbf{v}}_t^\dag \odot \hat{\mathbf{o}}_t^\dag))\\
    \hat{\mathbf{v}}_{t+1}=g(\mathbf{V}\mathbf{V}^\dag( \mathbf{z} \odot \hat{\mathbf{h}}_t^\dag \odot \hat{\mathbf{o}}_t^\dag))\\
    \hat{\mathbf{o}}_{t+1}=g(\mathbf{O}\mathbf{O}^\dag( \mathbf{z} \odot \hat{\mathbf{h}}_t^\dag \odot \hat{\mathbf{v}}_t^\dag)) 
\end{split}
\end{equation}

\noindent where $g(\cdot)$ is a function that normalizes complex amplitudes to 1, $\dag$ denotes complex conjugate (which in this case is an inverse of the binding operation), and the initial states, $\hat{\mathbf{h}}_0$, $\hat{\mathbf{v}}_0$ and $\hat{\mathbf{o}}_0$ are chosen randomly. We find that random initializations are conceptually useful because they provide a ``random seed'' to the otherwise deterministic dynamics, and, thus, they lead to the interpretation of the resonator network as a Monte Carlo algorithm. Finally, we use the asynchronous updates to factors as described in \cite{kent2020resonator}, as we find that this convention results in faster convergence. 

\section{Results}
To show the benefits of convolutional sparse coding, we compare our approach to a resonator network that instead uses image pixel values directly encoded into an HD/VSA vector. We find that convolutional sparse coding improves the performance of resonator networks in multiple ways, including:
\begin{itemize}
    \item \textit{higher accuracy} in factorization on single trials of the resonator;
    \item \textit{faster convergence} time to the correct solutions;
    \item improved factorization of \textit{multiple} objects, and
    \item higher \textit{confidence }in its final outputs; indicating better pattern separation.
\end{itemize}

We demonstrate that these trends hold on multiple datasets, focusing on two cases (``Random Bars'' and ``Translated MNIST''). Further experiments on a third dataset (``Letters'') are discussed in the Appendix.

\subsection{Random Bars}

\begin{figure*}
    \centering
    \includegraphics[width=\textwidth]{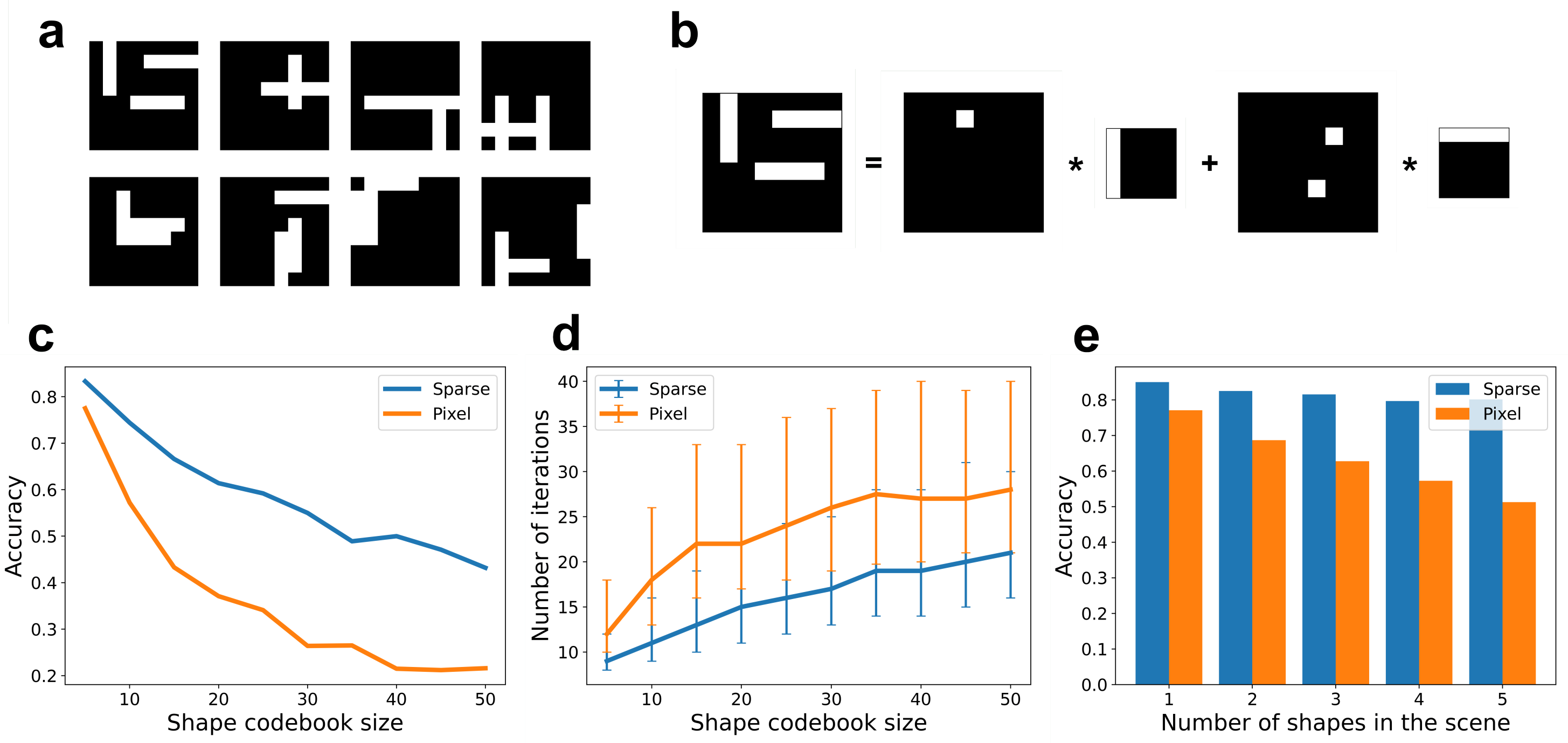}
    \caption{Results from the Random Bars dataset. a) Example objects from the Random Bars dataset. Each object consists of white horizontal and vertical bars that are placed at random positions. b) Ground truth convolutional sparse code, the two basis functions are white horizontal and vertical bars. c) Single trial accuracy of resonator network inference for a varying degree of items in the shape codebook, given a fixed dimension of $D=2500$. The sparse representation consistently outperforms the pixel encoding. d) The resonator network also converges in fewer iterations when using sparse representations. Solid lines report median number of iterations, with intervals reporting 25th and 75th percentiles. e) Sparse coding helps to improve decoding of multiple shapes; multi-object accuracy is higher and decays slower for increasing numbers of present shapes.}
    \label{fig:bars dataset}
\end{figure*}

First, to demonstrate the usefulness of sparse representations, we construct a custom dataset for which we have defined the ground truth basis functions $\{ \mathbf{\phi}_j \}$ for the convolutional sparse code representations. We fix two basis functions, a white vertical bar and a white horizontal bar and randomly place them in an $8 \times 8$ black square (Fig.~\ref{fig:bars dataset}a). This allows testing the resonator network on an arbitrarily large number of shapes, as well as using ground truth convolutional sparse code representations shown in Fig.~\ref{fig:bars dataset}b. This dataset is challenging to factorize because the shapes are composed of the same parts and have some overlap when translated. These structural similarities introduce local minima for the resonator network, in which the resulting factors explain only some of the pixels in the image.

To test how resonator network performance scales with the problem size, we study the impact of the number of entries in the shape codebook, $\mathbf{O}$.
Increasing the number of entries makes the total search space larger and makes it more difficult for the resonator network to converge, due to cross-talk noise between pseudo-orthogonal codebook vectors. For each size $K$ of the shape codebook (from $5$ to $50$), we randomly pick $K$ shapes from the bars dataset (Fig.~\ref{fig:bars dataset}a). We randomly choose one of them and add them at a random position to a $100 \times 100$ empty scene. We test our method and compare it to a method using the pixel values for HD/VSA encoding (similar to \cite{renner2022neuromorphic}):

\begin{equation}
    \mathbf{z}_{\text{pix}}(\mathbf{I})=\sum_{x,y} \mathbf{I}(x,y) \cdot \mathbf{h}(x) \odot \mathbf{v}(y),
\end{equation}

\noindent where $\mathbf{I}(x,y)$ is the pixel value of image $\mathbf{I}$ at position $(x,y)$.

We consider the factorization correct if the shape and position match the image, and incorrect otherwise. We set the maximum number of iterations for the resonator network to $100$, the HD/VSA dimension $D$ to $2{,}500$ and the resonator network is considered to have converged when the difference between two consecutive output vectors is smaller than $0.01$. We repeat this experiment $1{,}000$ times for each codebook size and report the accuracy and the average number of iterations in Figs.~\ref{fig:bars dataset}c and d, respectively. We observe that the convolutional sparse code representations are more effective than the pixel encoding, both in terms of accuracy and convergence speed.

We also evaluate the capability of the resonator network to factorize scenes with multiple shapes (Fig.~\ref{fig:bars dataset}e). This is achieved by allowing the resonator network to factorize one shape at a time and progressively explain away its estimates \cite{Frady2021Factorization}. We construct images with $m \in \{1,2,3,4,5\}$ shapes from the bars dataset, randomly placed in a $100 \times 100$ black scene. We encode the scene using the pixel and the sparse code representations and run the resonator network $m$ times. At each run, we subtract the output of the resonator network from the HD/VSA vector of the scene, proportionally to its similarity to the scene's representation. For run $r$, we denote $\mathbf{z}_r$ as the input of the resonator network and $\hat{\mathbf{z}}_r = \hat{\mathbf{h}}_r \odot \hat{\mathbf{v}}_r \odot \hat{\mathbf{b}}_r$ as its output.

\begin{equation*}
    \mathbf{z}_{r+1} = \mathbf{z}_r - \frac{|\left \langle \hat{\mathbf{z}}_r, \mathbf{z}_r \right \rangle|}{D} \hat{\mathbf{z}}_r.
\end{equation*}

\noindent This corresponds to explaining away the shape already found from the scene, for the resonator network to converge to another solution. The proportionality coefficient $\frac{|\left \langle \hat{\mathbf{z}}_r,  \mathbf{z}_r \right \rangle|}{D}$ measures whether the resonator network converged to an actual object in the scene. If the right object is found at the right position, the coefficient will be close to one. If not, it will be close to zero. Because explaining away does not use ground truth information, when the wrong object is decoded it contributes additive noise to the HD/VSA vector, which makes the problem harder for the decoding of remaining objects. Finally, the results in Fig.~\ref{fig:bars dataset}e show that the sparse representation is more effective than the pixel encoding for scenes with multiple objects. For such scenes, we evaluate accuracy on a graded basis (e.g., a scene in which 4 of 5 object-position pairs are identified is evaluated at 80 percent). The more shapes there are in the scene, the greater the probability that the shapes have partial spatial overlap. We observe that in cases of overlap, the pixel encoding leads to incorrect factorization, while the sparse representation remains effective. Pixel encodings often fail because of the local minima induced by pixel-wise correlations in the data. However, sparse coding reduces these correlations by assigning overlapping features to distinct feature maps, enabling the resonator network to more easily distinguish them.

\subsection{Translated MNIST}\label{subsection: MNIST}

\begin{figure*}
    \centering
    \includegraphics[width=\linewidth]{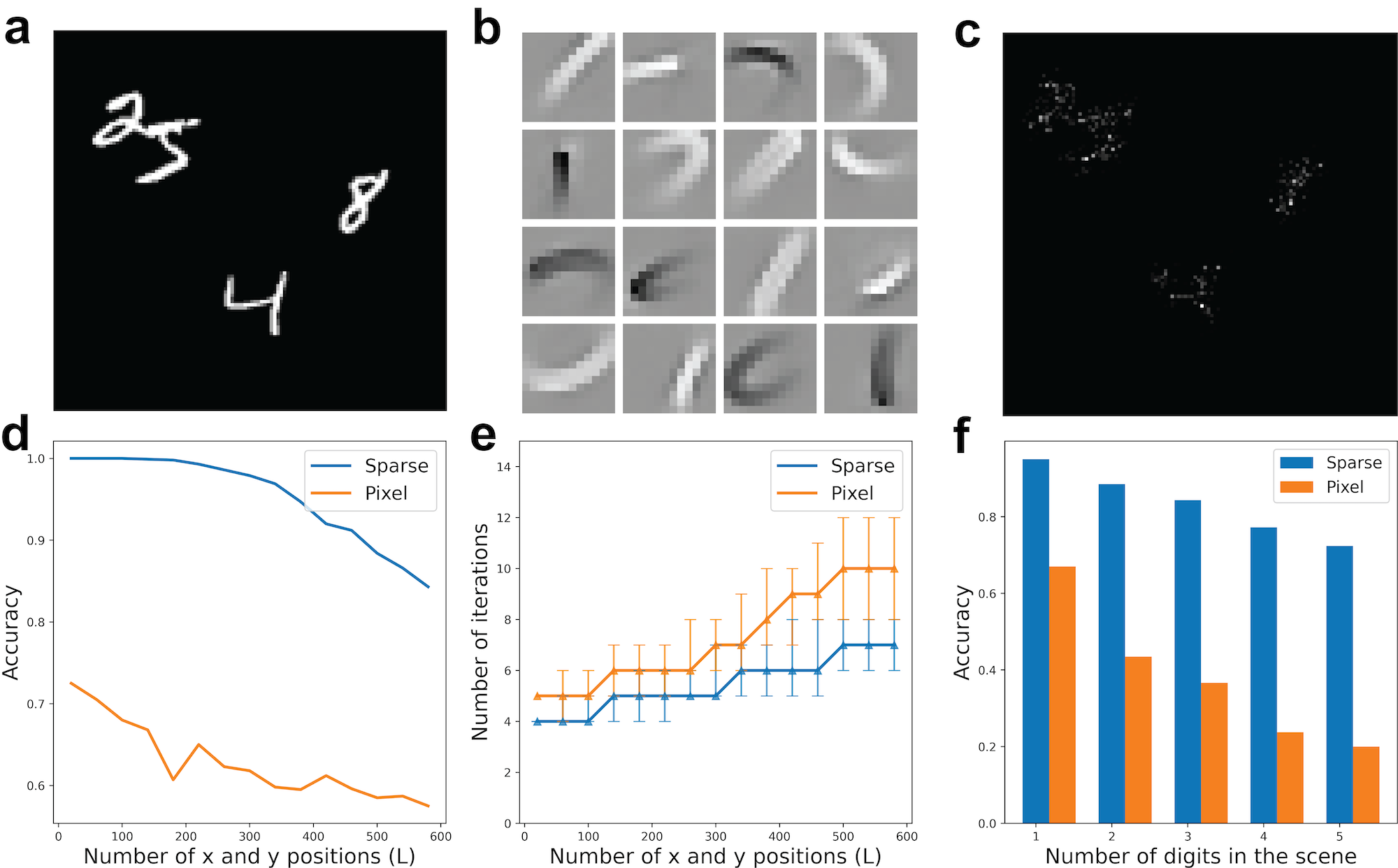}
    \caption{Results from the Translated MNIST dataset. a) An example scene composed of 4 handwritten digits from the Translated MNIST dataset. b) The learned set of basis functions $\{ \boldsymbol\phi_j \}$; each element resembles a localized stroke. c) Absolute values of the convolutional sparse code are averaged over different channels. A much lower fraction of spatial positions are active compared to the original image. d) Improvements of the sparse representation and the resonator network over the pixel encoding baseline, in  terms of accuracy on images of varying sizes. e) Sparse representations also reduce the required number of iterations for the resonator network to converge. Solid lines report median number of iterations, with intervals reporting 25th and 75th percentiles. f) Improvements of sparse representations when operating on scenes with multiple digits. These results are consistent with our findings on the Random Bars dataset.}
    \label{fig:mnist_results}
\end{figure*}

Next, we compare the results of our model on a more naturalistic dataset consisting of handwritten digits from the MNIST dataset. Unlike the Random Bars dataset, we do not have the ground truth dictionary and must, therefore, learn it from data. We optimize basis functions of the dictionary on a small subset of the MNIST dataset and then test the model inference on 10 prototypical digits, using the Python library SPORCO~\cite{wohlberg2017sporco}. More specifically, we train a set of $16$ basis functions ($\{ \phi_j \}$) of size $12 \times 12$ until convergence on $1{,}000$ samples of the MNIST training dataset. We find that the features learned resemble localized ``strokes'' that resemble parts of digits (Fig.~\ref{fig:mnist_results}b). Using these features, we can generate a sparse representation of an image with multiple objects (for an example, see Figs.~\ref{fig:mnist_results}a and c).

To evaluate how performance scales, we test the accuracy of factorization with the resonator network using a set of 10 prototypical digits over an increasing range of vertical and horizontal positions. Thus, the the search space is of size $10L^2$, where $L$ is the dimension of the image. Here, we set the maximum number of iterations for the resonator network to $20$, and the resonator network is considered to have converged when the difference between two consecutive output vectors is smaller than $0.05$. We use a fixed dimension ($D=10{,}000$) to demonstrate scaling as the resonator network factorization problem becomes more difficult; notably, for many values of $L$ this is smaller than the feature maps for sparse coding (which are of total size $16L^2$). We observe that for larger dimensions, the accuracy across all conditions improves, consistent with previous simulations with resonator networks~\cite{kent2020resonator}.

Consistent with our findings on the Random Bars dataset, we find that sparse representations consistently outperform pixel encodings. Notably, the single-trial accuracy remains near perfect for sparse representations even for hundreds of different possible positions (Fig.~\ref{fig:mnist_results}d). Across all settings, we also observe that the average number of iterations increases as the search space increases, with faster convergence of the resonator network when using sparse representations (Fig.~\ref{fig:mnist_results}e).

Finally, we evaluate the accuracy when factorizing multiple digits with the same explaining away procedure described for the Random Bars dataset. We find that the resonator network operating on sparse representations retains its advantage in terms of accurately finding digits in the scene (Fig.~\ref{fig:mnist_results}f).

\subsection{Confidence as an early-stopping criterion}
\label{sec:conf}

\begin{figure*}
    \centering
    \includegraphics[width=\textwidth]{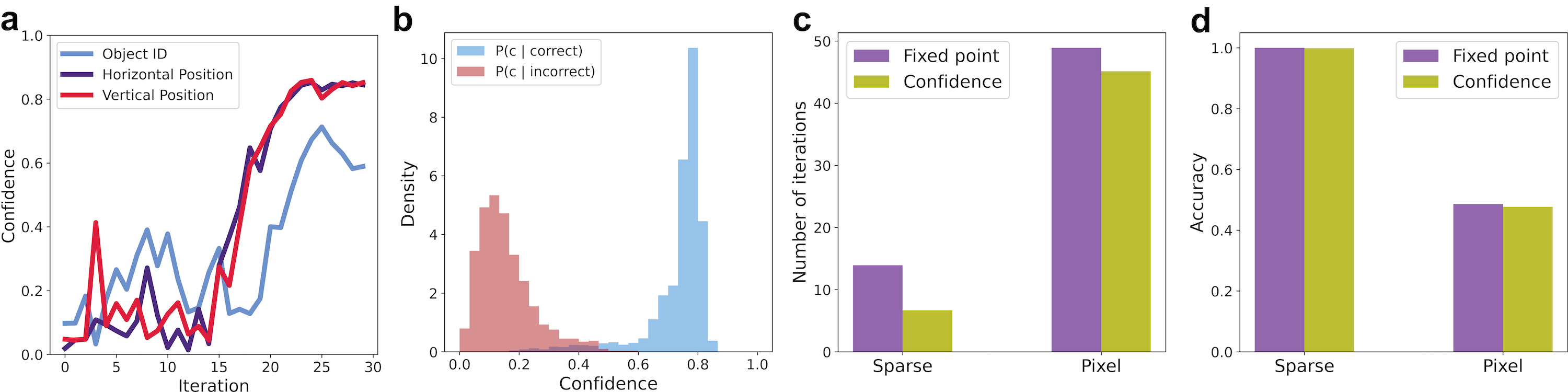}
     \caption{Confidence metrics, tested on the Translated MNIST dataset. a) Example evolution of confidence over the course of resonator network dynamics. Early stopping when all confidence values cross a threshold. b) Distribution of confidence values after convergence in the case when the resonator network is correct and incorrect. c) The confidence-based stopping criterion converges in fewer iterations compared to the fixed point stopping criterion (using thresholds of $0.6$ for sparse and $0.3$ for pixel). d) Using the thresholds in (c) are sufficient to achieve the same level of accuracy as with the fixed point criterion.}
    \label{fig:confidence}
\end{figure*}

Furthermore, we formulate a new confidence-based metric for measuring progress in resonator networks. We show that confidence scores are simple to compute, enable correct early stopping of resonator network dynamics, and are better taken advantage of by sparse coding resonator networks.

To measure the confidence,~\cite{chuang2020dynamic,kleyko2023efficient} introduce a confidence metric $c$, which is the mean of the difference between the two best candidates for the factorization in each codebook. Consider the codebook $\mathbf{P}$ ($\mathbf{H}, \mathbf{V}$ or $\mathbf{O}$) and the HD/VSA vector $\mathbf{\hat{p}_t}$ ($\mathbf{\hat{h}_t}, \mathbf{\hat{v}_t}$ or $\mathbf{\hat{o}_t}$) that the resonator network guesses at iteration $t$ for codebook $\mathbf{P}$. We compute the absolute values of inner products between each codebook vector in $\mathbf{P}$ and $\hat{p}_t$, denoting $a$ and $b$ as the highest and second highest resulting values, respectively. Then the normalized confidence metric is defined as:

\begin{equation}
    c(\mathbf{P},\mathbf{\hat{p}_t})=\frac{a-b}{a} \in [0,1].
\end{equation}

\noindent

Fig.~\ref{fig:confidence}a demonstrates an example of how confidence values change as the resonator network dynamics progress. We observe empirically  that many runs involve two stages: an early ``exploration'' stage in which confidence values remain low, and a ``confirmation'' stage in which confidence values progressively increase until convergence. Though the stages vary in their relative proportion, the key insight is that during the second stage, the estimate of the resonator network does not change, indicating that early stopping can save computations.

We also observe that the confidence metric can be a strong predictor of whether the resonator network has converged accurately or not. Fig.~\ref{fig:confidence}b shows the mean confidence values (averaged across codebooks) for the final resonator network's state (using sparse representations) for a fixed dimension and search space ($D=5000$, $L=250$). In this instance, the distributions are well-separated by a confidence threshold, also supporting confidence as an appropriate stopping criterion.

Therefore, we compare a confidence-based criterion for convergence to the fixed-point criterion introduced earlier. To evaluate the effectiveness of the confidence-based criterion, we evaluate the resonator network's performance on the Translated MNIST dataset with $D=5{,}000$, $L=120$, observing performance over $1{,}000$ trials. We find that under these conditions, confidence thresholds of $0.6$ for sparse representations and $0.3$ for pixel encodings are sufficient to observe a decreased number of iterations (Fig.~\ref{fig:confidence}c) without any sacrifice in accuracy (Fig.~\ref{fig:confidence}d). Notably, the mean confidence scores for correct resonator network factorizations is lower for pixel encodings, due to correlations between digit templates. Even though the threshold for confidence is higher in the sparse case, the confidence-based criterion results in faster convergence.

\section{Discussion}

\subsection{Summary of the study}

We presented a principled approach for disentanglement of visual scenes based on two key ideas: convolutional sparse coding and resonator networks. Though both ideas have been proposed previously, and independently, we show how convolutional sparse coding provides an equivariant, data-adaptive encoding scheme that can be converted to a high-dimensional vector suitable for factorization by a resonator network. We show that this formulation improves over pixel-based encoding schemes on several metrics: accuracy, convergence time, and confidence, even for scenes with multiple objects.

Prior studies relating HD/VSA to visual scene analysis can be clustered into two strategies. The first strategy~\cite{FradyResonator2020,hersche2023neuro} treats images as symbolic relations of objects, defines the correct mapping from images to high-dimensional vectors \textit{a priori} and trains a neural network to learn that mapping via supervised learning. By contrast, we adopt an unsupervised learning approach to more compactly describe latent structure in images and use the VFA encoding strategy to handle continuous valued-variables in images. Our approach is, therefore, similar to the second strategy used by~\cite{renner2022neuromorphic}, but instead encoding the coefficients resulting from sparse coding, rather than the pixel values directly. Our results demonstrate that this change makes a significant difference in the performance of the resonator network's factorization.

Conversely, prior work on convolutional sparse coding has focused on applications such as image reconstruction or super-resolution (e.g.,~\cite{gu2015convolutional}), with applications to disentanglement problems largely underexplored. Of models working on similar tasks (such as segmentation), these tend to be in the setting where brute force factorization is tractable, leaving open the possibility for developing more efficient algorithms.

The main advantages of our approach are its compositionality and its transparency. The model is explicitly compositional because it treats all images as constructed from a relatively small set of codebook entries and operations (binding and bundling). Compositionality enables high expressivity with relatively few codebook entries, as well as easy generalization to new instances by adding new entries to the object codebook. Transparency comes both from the fact that sparse coding makes the structure of images compact yet explicit, and from the fact that at each time step, the resonator network's estimates can be readily understood by comparing inner product values with codebooks. The confidence metric that we develop utilizes this correspondence to allow for early stopping in resonator networks without losses in the accuracy.

\subsection{Related work}

\textit{Relations to existing HD/VSA models.} Though HD/VSA were initially formulated as models of symbolic reasoning with random, high-dimensional vectors~\cite{kleykosurveyvsa2021part1, kleykosurveyvsa2021part2}, a wave of recent work has focused on extending this framework to continuous variables and functions~\cite{KomerNavigation2020,fradyfunctionsnice2022, FurlongProbabilistic2023}. These connections can be made rigorous by thinking of them as randomized kernel approximation~\cite{rahimi2007random,thomas2021theoretical}. Here, we show that this framework extends naturally to convolutional sparse coding, providing a general way for image representation.

\textit{Connections to computational neuroscience.} Our two-stage model also has structural similarities to several models proposed in computer vision. In particular, it falls under the class of models whose generative models express multilinear functions~\cite{olshausen2007bilinear}. Perhaps this should not be surprising, given that these functions enable tractable inference yet are expressive enough for a large class of operations.

An interesting line of work has focused on learning both basis functions via sparse coding and common classes of transformations (e.g., rotation and translation). In particular,~\cite{chau2023disentangling} learn both object templates and irreducible representations of commutative, compact, connected groups (expressing transformations). Remarkably, this generative model is equivalent to our resonator network circuit but uses a different strategy to infer latent representations for a given image. Because the inference strategy of their method is effectively brute force in the transformation space, an interesting direction for future work is to understand how the resonator network could improve inference.

We also observe the similarity to the Map Seeking Circuit (MSC)~\cite{arathorn2002map}, which provides an alternative approach to solving visual scene factorization problems using the principle of search-in-superposition. Previous work on resonator networks has remarked on the similarities to MSC, most notably in~\cite{kent2020resonator}. Yet there are also clear differences in how the dynamics for each system are defined. At a high level, the resonator network aims at self-consistency between its different inputs to explain an image; whereas the MSC refines its estimate by \textit{culling} incorrect inputs. It is intriguing to speculate whether there are further emergent similarities in how the approaches solve factorization problems; our confidence metric suggests one quantitative basis for such a comparison.

\subsection{Future work}

Finally, we mention a few more extensions to our approach that seem promising. One direction is to apply our approach to more complex classes of transformations, for instance, by learning dictionaries of basis functions for convolutional sparse coding and common transformations in time-varying natural scenes \cite{sohl2010unsupervised,cadieu2012learning}. A second direction is to incorporate learnable modules within the resonator network, to allow the objects to be representable for classes of positions. A third direction is to explore extensions suggested by other work on resonator networks, such as efficiency gains from using residue number systems \cite{kymn2023computing}, nonlinearities in the resonator network dynamics \cite{langenegger2023memory}, and log-polar coordinate transformations to handle cases such as rotation and scaling~\cite{renner2022neuromorphic}.

A second area for future work is to design implementations of our approach on neuromorphic hardware. Sparse coding has been implemented in spiking neural networks~\cite{zylberberg2011sparse} and in neuromorphic hardware~\cite{chavez2023spiking}, as have complex-valued attractor neural networks like the resonator network~\cite{frady2019robust,renner2022neuromorphic}. This work complements recent efforts on in-memory computing implementations for HD/VSA~\cite{langenegger2023memory}, fulfilling promises of HD/VSA as a paradigm for algorithmic-level abstraction in emerging hardware~\cite{kleyko2022vector}.

{\footnotesize
\bibliographystyle{ieeetr}
\bibliography{references}
}

\begin{figure*}[h]
    \centering
    \includegraphics[width=\textwidth]{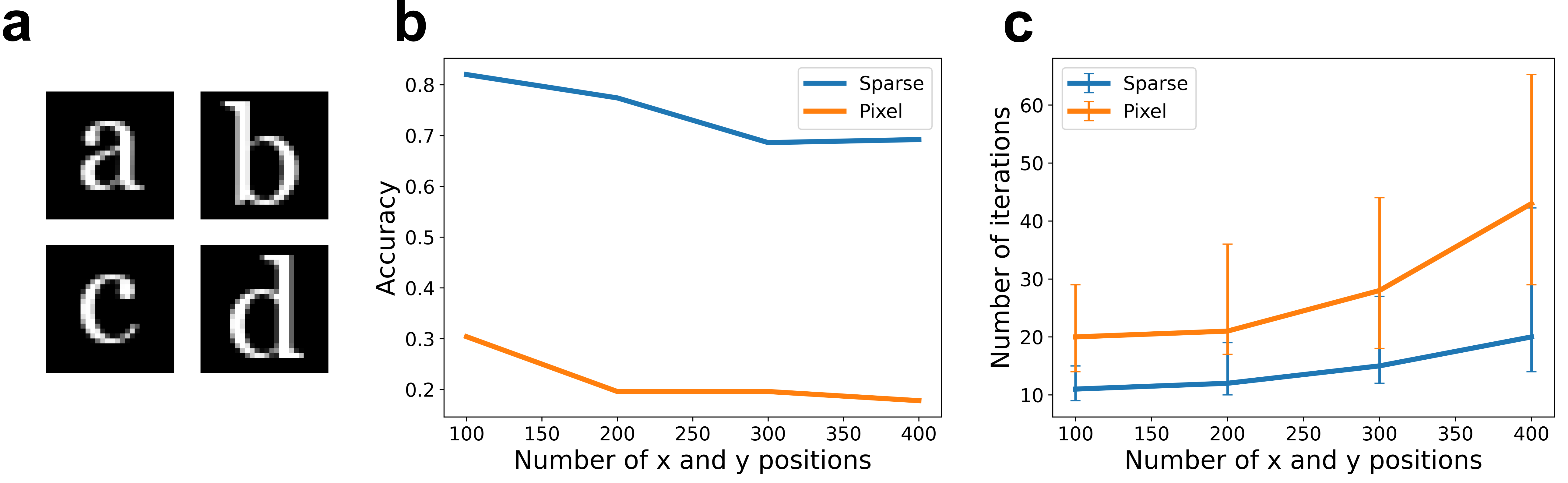}
    \caption{Results on the Letters dataset. a) Four example images (out of $26$) from the Letters dataset. b) Resonator network factorization is substantially more accurate when working with sparse code representations as opposed to the pixel encodings, for visual scenes of varying sizes. c) The convergence of the resonator network is also faster when factorizing sparse code representations. Data points report medians, and bars indicate $25$th and $75$th percentiles. These results are consistent with experiments on other datasets (see Figs.~\ref{fig:bars dataset} and \ref{fig:mnist_results}).}
    \label{fig:letters dataset}
\end{figure*}

\begin{figure*}[h]
    \centering
    \includegraphics[width=0.75\textwidth]{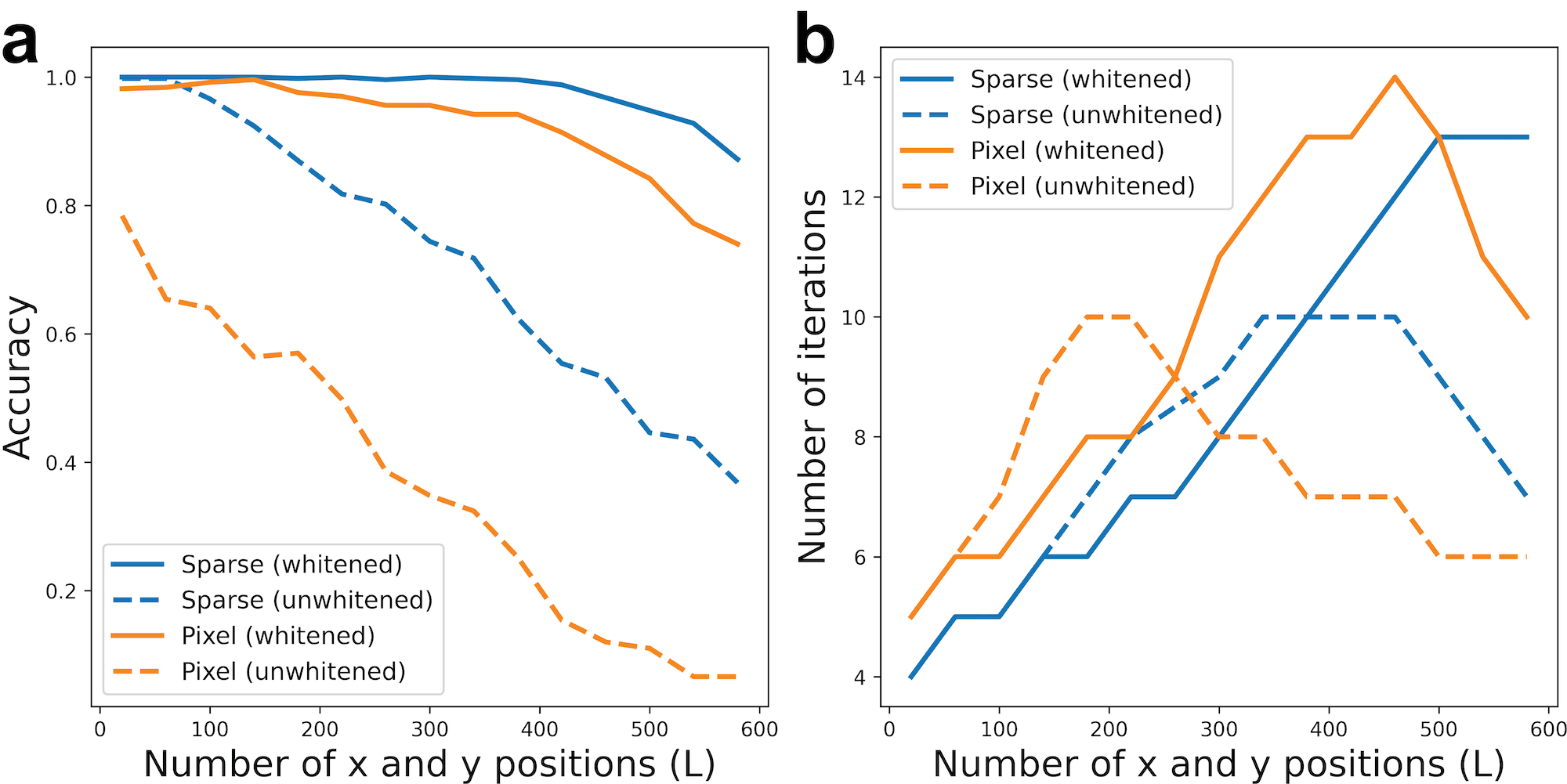}
    \caption{A comparison of the resonator network's performance on the Translated MNIST dataset when taking whitening into account ($D=2{,}500$, which is lower than for results shown in Fig.~\ref{fig:mnist_results}). a) Whitening improves the performance of both pixel encodings and sparse representations. b) A comparison of median number of iterations for both whitened and unwhitened vectors. When whitening transformations are applied, sparse representations still converge in fewer iterations compared to pixel encodings. The eventual decline in the number of iterations for low accuracy regimes indicates that, in these situations, the resonator network is often converging to local minima.}
    \label{fig:mnist_white}
\end{figure*}

\section{Acknowledgments}

The work of CJK was supported by the Department of Defense (DoD) through the National Defense Science \& Engineering Graduate (NDSEG) Fellowship Program. The work of SM was carried out as part of the ARPE program of ENS Paris-Saclay. The work of DK and BAO was supported in part by Intel’s THWAI program. The work of CJK and BAO was supported by the Center for the Co-Design of Cognitive Systems (CoCoSys), one of seven centers in JUMP 2.0, a Semiconductor Research Corporation (SRC) program sponsored by DARPA. 
DK has received funding from the European Union’s Horizon 2020 research and innovation programme under the Marie Sklodowska-Curie grant agreement No 839179.

\appendix

\subsection{Letters dataset}

We test our approach on the third dataset, composed of $26$ images of Latin alphabet letters in Georgia font (examples are shown in Fig.~\ref{fig:letters dataset}a). This dataset has more structure than the handwritten MNIST dataset, which makes it more difficult for the resonator network. For example, most letters have the same straight lines, which creates local minima for the resonator network. We do the same experiments as for the translated MNIST dataset in Section \ref{subsection: MNIST}, but there are $26$ image classes rather than $10$. The HD/VSA dimension is $5{,}000$. Consistent with our results on the other datasets, we find that convolutional sparse coding improves both accuracy (Fig.~\ref{fig:letters dataset}b) and convergence time (Fig.~\ref{fig:letters dataset}c). 

\subsection{Experiments with whitening}

Whitening is a common pre-processing step when applying HD/VSA-based algorithms to datasets~\cite{renner2022neuromorphic, frady2019robust}. The motivation is that algorithms such as the resonator network work best when dealing with vectors that are pseudo-orthogonal, in order to minimize cross-talk noise when working in superposition. Whitening rotates a data matrix $\mathbf{X}$ by a linear transformation $\mathbf{W}$ such that the new matrix $\mathbf{WX}$ has the identity matrix as its covariance; one way of generating $\mathbf{W}$ is based on the Singular Value Decomposition (SVD) of $\mathbf{X}$.

Although whitening is a useful pre-processing step, it is not immediately clear whether it is appropriate for our problem setting. It is straightforward enough to whiten the templates of objects, but less obvious how to also apply this whitening to new scenes with different dimensions, and with shifted and superimposed objects. Whitening occurs relative to a standard frame of reference, but guarantees of orthogonality no longer hold when such templates are shifted. Previous efforts invoking whitening made the assumption that visual scenes contained objects composed of sums of whitened templates, which is importantly distinct from whitened sums of templates.

Fortunately, our main results demonstrate that when working with convolutional sparse coding, it is possible to improve factorization without resolving this question. Notably, the sparse code representations are decorrelated enough to reach nearly perfect accuracy, whereas in datasets such as Translated MNIST, the correlations can often cause the resonator network to converge to local minima.

Finally, we find that if we perform whitening in either the pixel space or the sparse coding coefficient space, these advantages stack with the benefits of sparse representations. In the pixel space, this corresponds to SVD whitening the image templates, for sparse code representations, we apply this to the convolutional feature maps. We report results for the MNIST dataset, observing increased accuracy (Fig.~\ref{fig:mnist_white}a) and often less iterations (Fig.~\ref{fig:mnist_white}b) with whitening. For these results, we use a much lower dimension ($D=2{,}500$) than for experiments shown in Fig.~\ref{fig:mnist_results}, because this lower dimension shows a range in which accuracy with whitened dictionaries tends to decay. Intriguingly, the number of iterations begins to decrease as the resonator network becomes more inaccurate at inference; this has been previously reported in regimes where there are too many codebook vectors for a fixed dimension \cite{kent2020resonator}.

\end{document}